\documentclass[journal]{IEEEtran}
\ifCLASSINFOpdf
\else
\fi
\usepackage{amsmath,epsfig,url,comment}
\usepackage{multirow}
\usepackage{booktabs} 
\usepackage{siunitx}   
\usepackage{placeins}  
\usepackage[colorlinks=true, linkcolor=black, urlcolor=blue, citecolor=black]{hyperref}
\usepackage{array}

\begin{document}
%
\title{NeRF-Accelerated Ecological Monitoring in Mixed-Evergreen Redwood Forest}
%
%
%

\author{Adam~Korycki~\IEEEmembership{Member,~IEEE,}
        Cory~Yeaton,
        Gregory S. Gilbert,\\
        Colleen Josephson~\IEEEmembership{Member,~IEEE,}
        Steve McGuire~\IEEEmembership{Member,~IEEE}
\thanks{A. Korycki, C. Josephson, and S. McGuire are with the Department of Electrical and Computer Engineering, UC Santa Cruz, CA 95060 USA}
\thanks{C. Yeaton is with the Department of Ecology and Evolutionary Biology, UC Santa Cruz, CA 95060 USA}
\thanks{G. Gilbert is with the Department of Environmental Studies, UC Santa Cruz, CA 95060 USA}}

\maketitle

\begin{abstract}
Forest mapping provides critical observational data needed to understand the dynamics of forest environments. Notably, tree diameter at breast height (DBH) is a metric used to estimate forest biomass and carbon dioxide (CO$_2$) sequestration. Manual methods of forest mapping are labor intensive and time consuming, a bottleneck for large-scale mapping efforts. Automated mapping relies on acquiring dense forest reconstructions, typically in the form of point clouds. Terrestrial laser scanning (TLS) and mobile laser scanning (MLS) generate point clouds using expensive LiDAR sensing, and have been used successfully to estimate tree diameter. Neural radiance fields (NeRFs) are an emergent technology enabling photorealistic, vision-based reconstruction by training a neural network on a sparse set of input views. In this paper, we present a comparison of MLS and NeRF forest reconstructions for the purpose of trunk diameter estimation in a mixed-evergreen Redwood forest. In addition, we propose an improved DBH-estimation method using convex-hull modeling. Using this approach, we achieved 1.68 cm RMSE, which consistently outperformed standard cylinder modeling approaches. Our code contributions and forest datasets are freely available at \href{https://github.com/harelab-ucsc/RedwoodNeRF}{https://github.com/harelab-ucsc/RedwoodNeRF}. 
\end{abstract}


%
\IEEEpeerreviewmaketitle

\section{Introduction}
\label{sec:intro}
Forests are the Earth's largest terrestrial  carbon store, holding more than three decades worth of global CO$_{2}$ emissions~\cite{EPA} and consuming a quarter of new anthropogenic emissions~\cite{davies2021forestgeo}. Pressingly, climatic trends are revealing grave uncertainty for long-term stability. According to U.S. Forest Service aerial surveys, over 200 million trees died in California since 2010, with 62 million dead in 2016 alone~\cite{USFS2016}. The warming climate and the consequence of longer, more severe drought cycles is the primary culprit of this mass die-off. Significant numbers of dead and dying trees dramatically increases the risk of wildfires; these counts do not include tree deaths caused by wildfires, which adds hundreds of millions to the toll.

Forest management is a recognized, cost-effective approach to mitigating the effects of the climate crisis~\cite{TreePlanting}. Global carbon accounting, which is a crucial contribution to informed climate change policies, relies on large-scale forest surveys. Tree diameter at breast height (DBH) is a primary data point used in ecological monitoring and carbon accounting efforts; the conventional means of determining DBH relies on a human forester with a measuring tape.
\begin{figure}[t!]
    \centering
    \includegraphics[width=8cm]{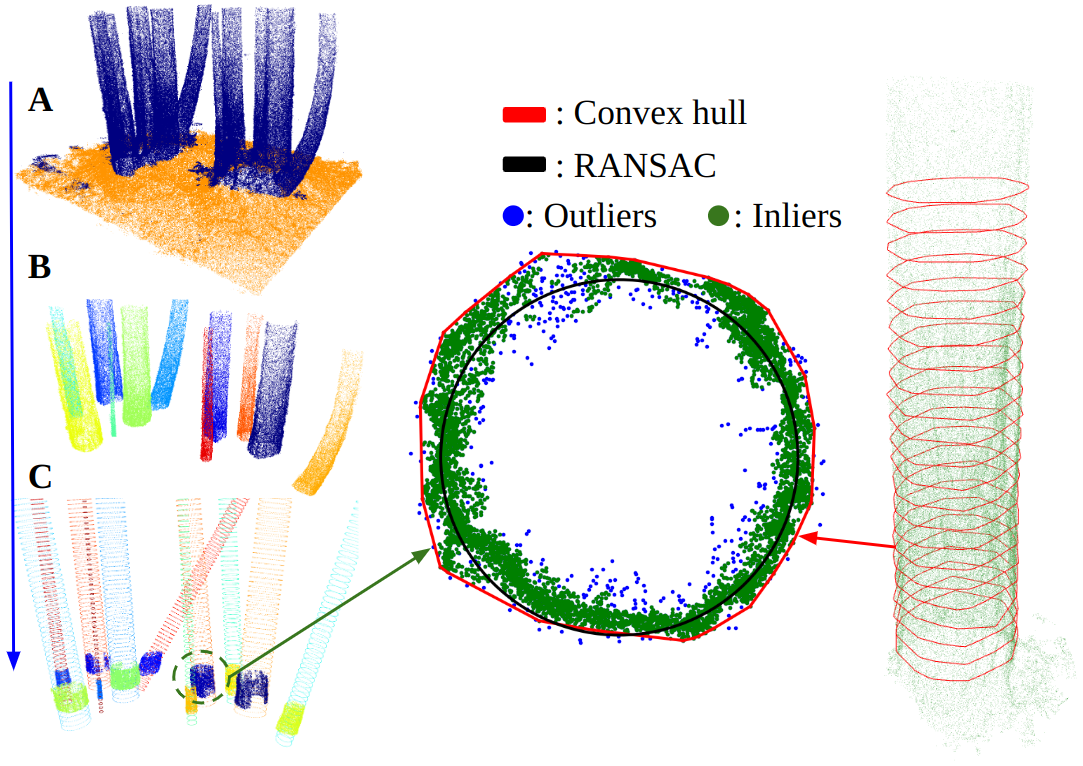}
    \caption{TreeTool process applied to a forest NeRF reconstruction. (A) ground segmentation, (B) trunk segmentation, and (C) trunk modeling. Our tree modeling approach considers trees as stacks of convex-hull slices which outperformed other approaches by 3-4$\times$ in terms of DBH estimation accuracy.}
    \label{fig:treetool}
\end{figure}
Accurate, automated methods of DBH estimation could drastically reduce the time and effort needed to perform surveys, opening doors for large-scale mapping efforts. Three dimensional reconstruction is the task of digitally representing real world settings, typically in the form of a point cloud. Metric reconstruction is a subclass which preserves true scale in the recovered geometry and affords the ability to indirectly measure scene features (e.g. volume, length). Three dimensional (3D) reconstruction using terrestrial laser scanning (TLS) to map forest inventories offers the potential for rapid ecological assessment, but these systems cost 80,000-120,000 USD, and additionally require a skilled operator. Survey-grade terrestrial laser scanners can provide centimeter-level diameter estimation in forests with uniform tree structures and even terrain~\cite{LIANG}.  The main technical issue faced by TLS methods is tree occlusions that require stitching many scans together from different spatial locations, a critical step in recovering complex geometries of forests. Research toward using mobile robot platforms in combination with Simultaneous Localization and Mapping (SLAM) algorithms addresses this problem using optimized pose-graphs to align thousands or millions of LiDAR scans taken along a robot's trajectory. A 2017 paper~\cite{pierzchala2018mapping} using SLAM reports a best-case DBH estimation RMSE of 2.38 cm for well-represented trees. A 2024 study~\cite{freissmuth2024online} achieved 1.93 cm RMSE using a mixed Hugh-RANSAC trunk modeling approach. However, these solutions require expensive 3D LiDAR and inertial measurement unit (IMU) hardware (10,000-25,000 USD).

Recent advances in the fields of computer vision and deep learning offer a new paradigm for generating 3D reconstructions. Neural Radiance Fields (NeRFs)~\cite{nerf} are an emergent technology enabling the recovery of complex 3D geometry by training a neural network on conventional imagery. NeRFs are a remarkable advance over traditional photogrammetry, producing higher quality, photorealistic 3D reconstructions from sparser input imagery and at an accessible efficiency. Since 2020, a community of developers has contributed hundreds of methodological improvements, rapidly improving its performance and accessibility. The ability to export NeRFs as point clouds lends itself as an aggressive alternative to expensive LiDAR-based mapping. 

We present an evaluation of NeRF-based forest reconstruction for the task of DBH estimation of mixed-evergreen redwood forest located in Santa Cruz, California. This study compares NeRF reconstructions trained on conventional mobile phone imagery to LiDAR-inertial SLAM reconstructions sourced from a quadruped robot equipped with a custom multi-modal sensing platform. This paper also expands the viability and accuracy of TreeTool~\cite{treetool}, a Python toolkit for rendering DBH estimates from forest point clouds. Specifically, we propose a new set of features to support robust tree detection and accurate DBH estimation. Many studies have relied on cylinder-fitting to model trunk morphology. We observe a DBH underestimation trend with this method and propose an improved convex-hull approach. In summary, the contributions of this work are:

\begin{itemize}
    \item Quantitative field study evaluating the performance of NeRF-based forest reconstructions compared to LiDAR-inertial SLAM with regards to DBH estimation accuracy.
    \item Improved DBH estimation accuracy via a trunk modeling approach using convex-hull and density-based filtering methods.
    \item Open-source modeling code and forest datasets including SLAM and NeRF reconstructions of a mixed-evergreen Redwood forest.
\end{itemize}

\section{Theoretical background}
\label{sec:related_work}
\subsection{The SLAM approach}
The SLAM problem can be broken into two tasks: building a map of the environment and simultaneously estimating the robot's trajectory within that map. More specifically, given all sensor measurements $z_{1:T}$, all robot motion commands $u_{1:T}$, and an initial robot position $x_0$, estimate the posterior probability of the complete robot trajectory $x_{1:T}$ and a globally consistent map \textbf{m} of the explored environment~\cite{grisetti2010}:
\begin{equation}
p(x_{1:T},\boldsymbol{m}|z_{1:T},u_{1:T},x_0)  
\end{equation}

This Bayesian problem formulation benefits from the ability to fuse several sensing modalities together while accounting for individual sensor noise. The modern-day solution to SLAM formulates the problem as a factor graph~\cite{thrun} where each node represents a robot pose $x_i$ (relative 3D spatial location and orientation) and each edge represents a transformation $T_i$ between nodes. Edges also encapsulate loop-closure constraints. Loop-closure is the subprocess of identifying previously observed landmarks in order to correct for drift and error accumulation in the estimated trajectory. Based on the prior pose-graph, a global optimization is used to minimize the estimation error given the sensor measurements and loop-closure constraints. The resulting pose corrections are back propagated, resulting in reduced error of scan alignment and dense scene reconstruction.

\subsection{NeRF scene representation}
\label{sec:NeRF}
NeRF~\cite{nerf} is a current state-of-the-art solution to the problem of novel view synthesis. This problem involves generating an image of a 3D scene from a particular view when the only available information is images from other views. Two pivotal ingredients to the NeRF method are continuous volumetric representation and deep fully-connected network architecture. NeRFs inherit the exceptional photorealism and reconstruction fidelity of continuous representation at a fraction of the storage cost compared to discrete approaches using voxels or meshes~\cite{nerf}. The scene is modeled as a Multilayer Perceptron Network (MLP) which takes a 5D input vector composed of spatial location \textbf{X} = ($x, y, z$) and viewing angle \textbf{d} = ($\theta, \phi$) and learns the weights $\Theta$ to map each 5D coordinate to the corresponding 4D output vector of color \textbf{c} = (r,~g,~b) and volume density ($\sigma$). The 5D input space is sampled using ray tracing. The network architecture is two MLPs. The first learns only volume density based on input location. The second learns color based on location and viewing direction as well as $\sigma$. This enables multi-view consistency, critical for non-Lambertian lighting (lighting conditions that include high dynamic range across the scene). A positional encoding layer is used to better represent high-frequency color-density functions. Instead of performing volume rendering uniformly along the rays, the NeRF method uses hierarchical volume sampling to identify relevant regions of the scene and avoid excessive calculations in rendering free space.
\begin{figure}[h!]
    \centering
    \includegraphics[width=8cm]{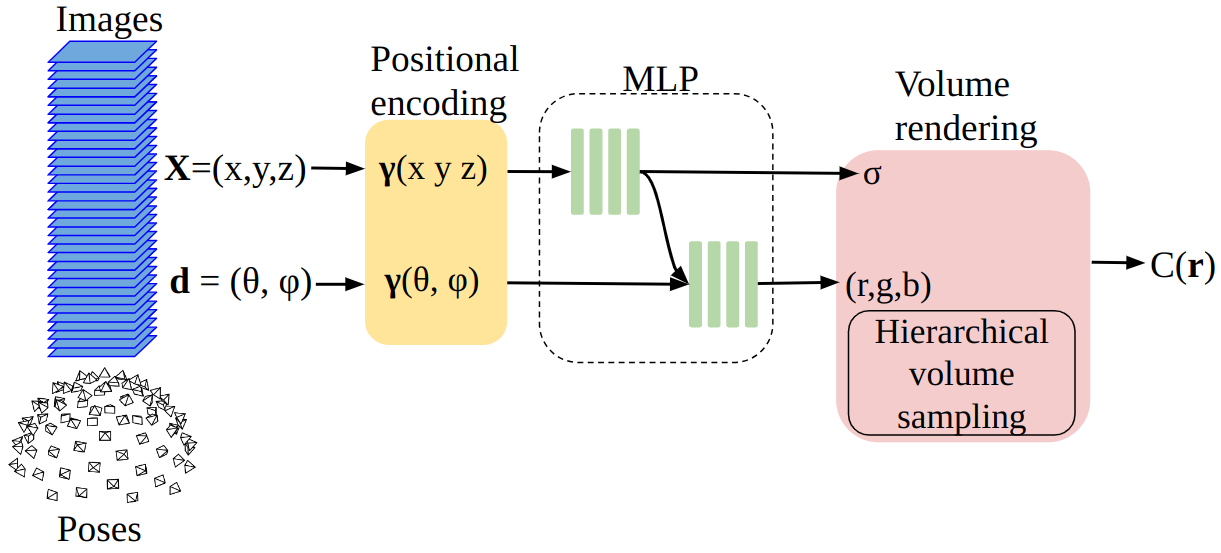}
    \caption{NeRF scene representation flow. Sparse images with corresponding poses are sampled using ray-tracing to generate 5D input vector comprised of location ($x, y, z$) and viewing direction ($\theta, \phi$). A cascaded MLP learns the weights to map this 5D vector to output color (r, g, b) and volume density $\sigma$. Volume rendering composites the learned rays to novel views.}
    \label{fig:nerf_pipeline}
\end{figure}

Scene reconstructions are rendered from the learned color-ray space by filtering low color-density regions to only leave surfaces. In order to extract and accurately measure geometric features, the image poses must be \textbf{metrically relevant} (in real-world distance units). Standard Structure from Motion (SfM) pipelines (such as COLMAP~\cite{schonberger2016structure}) are not able to maintain real-world scale since absolute depth is not available from monocular multi-view synthesis without additional information about the scene. Huang et al.~\cite{huang2024evaluating} use SfM poses to generate NeRF reconstructions of two trees. To cope with scale ambiguity, the authors derive a scale factor from a ground truth TLS reconstruction, which is impractical in many real-world cases where this external information is unavailable. The prevalent solution is to derive metrically-accurate camera poses from visual-inertial (VI) or LiDAR-inertial SLAM.

\section{Design and methods}
\label{sec:methods}
\subsection{Mobile laser scanning via LiDAR-inertial SLAM}
In order to perform SLAM-based reconstruction, we designed a robot based on the Unitree B1 quadruped platform. Terrain maneuverability was a prioritized feature to cope with rough, uneven forest terrain and complex obstacles. A custom built multi-modal sensor head is attached which includes LiDAR, stereo vision, inertial, and GNSS+RTK sensing modes. For online processing, the robot is equipped with an external x86 mini computer which includes a 4.5 GHz Core i7-1270pe CPU, 64 GB RAM, and 2 TB storage. The LiDAR is an Ouster OS0-128 with 90$^\circ$ vertical field-of-view and 128 horizontal channels. The IMU is an Inertialsense IMX5 capable of 1KHz output, and fused EKF attitude estimates. 

The computer runs ROS2 Iron for sensor recording and running LiDAR inertial odometry smoothing and mapping (LIOSAM)~\cite{liosam}. This software fuses LiDAR and IMU data together to create dense spatial reconstructions in real-time along with optimized pose estimations. LIOSAM tightly couples LiDAR and inertial data in a joint optimization using a factor-graph SLAM architecture. Through loop-closures factors, LIOSAM is able to achieve minimal drift in large exploration volumes~\cite{liosam}.

\begin{figure}[h!]
    \centering
    \includegraphics[width=8.8cm]{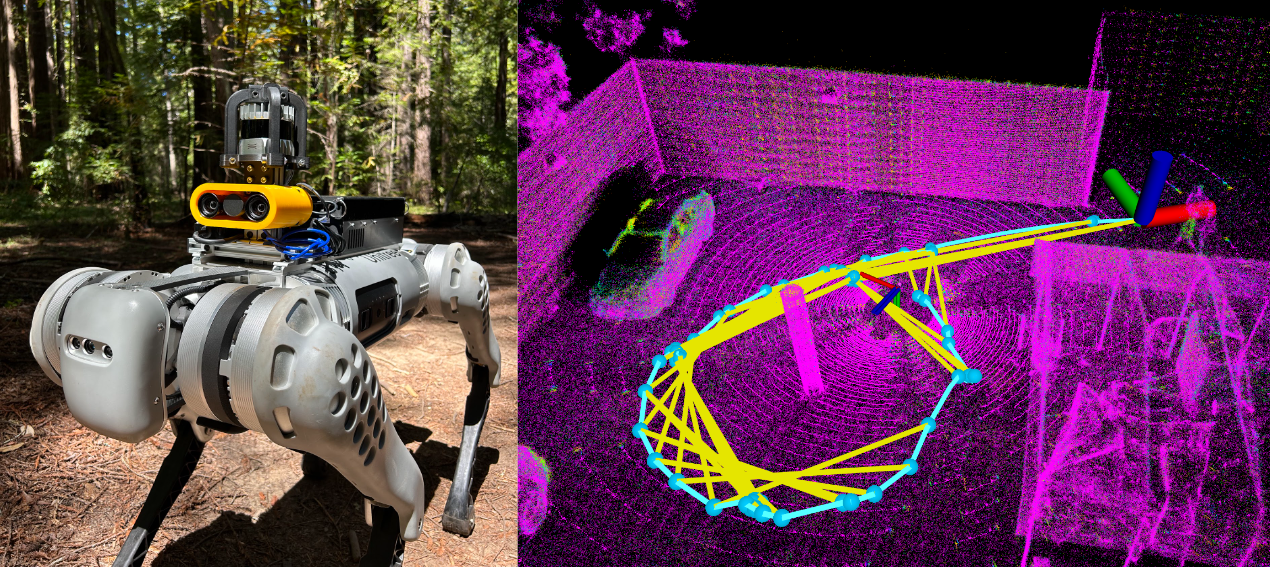}
    \caption{Quadruped robot creating a dense LiDAR-inertial reconstruction in a forest environment (left). LIOSAM visualization of estimated trajectory (torqouise), loop-closure events (yellow), and tightly aligned LiDAR scans (magenta).}
    \label{fig:mls_liosam}
\end{figure}

\subsection{NeRF reconstruction pipeline}
\subsubsection{Training data}
Metrically-relevant camera poses are needed in order to measure world features from NeRF reconstructions. While COLMAP can perform high quality vision-based reconstruction, it suffers scale ambiguity. The typical solution is to use VI or LiDAR-inertial SLAM pose estimation in which the metric information is derived from IMU measurements or from LiDAR range measurements.. In this study, we use an iOS application called NeRFCapture~\cite{nerfcapture} which uses Apple ARKit to provide camera poses in real time. ARKit uses VI odometry with multi-sensor fusion which lends itself as a good option for metric pose estimation. 
\subsubsection{Software implementation}
The base NeRF method discussed in Section ~\ref{sec:NeRF} has seen hundreds of proposed improvements over the years. Nerfacto~\cite{zhang2021nerfactor} is a method which draws from several other methods~\cite{instant-ngp, barron2021mip, wang2021nerf, martin2021nerf, verbin2022ref} and has been shown to work well in a variety of in-the-wild settings. For this reason, we chose the Nerfacto method for this study.

Nerfacto improves on the base method in a few key dimensions, the first of which is \textbf{pose refinement}. Error in image poses results in cloudy artifacts and loss of sharpness in the reconstructed scene. The Nerfacto method uses the back-propagated loss gradients to optimize the poses for each training iteration. Another improvement is in the ray-sampling. Rays of light are modeled as conical frustums and a \textbf{piece-wise sampling} step uniformly samples the rays up to a certain distance from the camera origin, and then samples subsequent sections of the conical ray at step sizes which increase with each sample. This allows for high-detail sampling of close portions of the scene, while efficiently sampling distant objects as well. The output is fed into a \textbf{proposal sampler} which consolidates sample locations to sections of the scene which contribute most to the final 3D scene render. The output of these sampling stages is fed into the Nerfacto field which incorporates \textbf{appearance embedding}, accounting for varying exposure among the training images. 

We used the nerfstudio~\cite{nerfstudio} API which makes training and exporting NeRF reconstructions extremely simple. Posed image data is copied to a remote desktop PC for training. This computer hosts a 3.8 GHz AMD 3960X CPU, 64 GB RAM, and 2 TB storage. The PC is also outfitted with two NVIDIA RTX-3070 graphics cards which aggregate to 16 GB of VRAM. The system runs Ubuntu 22.04 with CUDA-11.8 to interface with GPU hardware.

\begin{figure*}[!ht]
    \centering
    \includegraphics[width=0.9\textwidth]{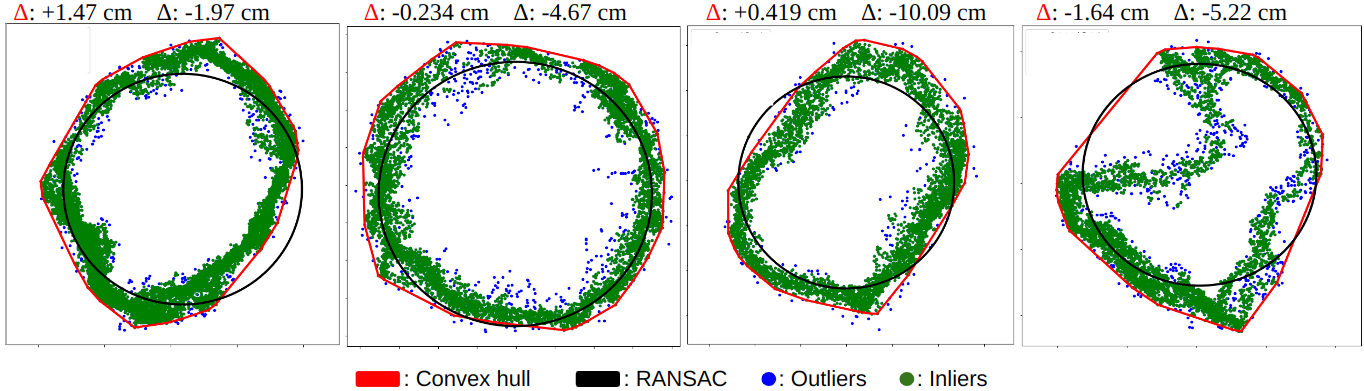}
    \caption{Four comparisons of RANSAC and convex-hull modeling approaches. Deltas between manual DBH and each modeling approach are provided on the top line. RANSAC cylinder modeling consistently under-fits well-represented trunk projections. Convex hull DBH estimation outperformed RANSAC by 3-4$\times$.}
    \label{fig:convex_hulls}
\end{figure*}
\subsection{Tree segmentation and modeling}
\subsubsection{TreeTool framework}
To process forest reconstructions and estimate tree DBH, we use TreeTool~\cite{treetool}, a Python library built on Point Data Abstraction Library (PDAL) and Point Cloud Library Python (pclpy). TreeTool breaks the process down into three distinct steps covering filtering, detection, and modeling stages. 

The goal of the filtering stage is to remove all non-trunk points, mainly the ground and foliage. The ground is segmented using an improved simple morphological filter (SMRF) proposed by Pingel et al.~\cite{pingel2013improved}. This technique uses image-inpainting to accurately model complex uneven terrain. Once the ground points are removed, TreeTool uses a surface-normal filter to remove foliage points. This is based on their observation that the surfaces created by trunk points have horizontal normals. TreeTool also filters non-trunk points by considering the curvature of surfaces. Curvature is interpreted as the percentage of information held by the eigenvalue associated with the normal vector. High-curvature points are discarded, which removes foliage.

The detection stage groups the filtered trunk points into individual stem sections. TreeTool uses Euclidean-distance to perform nearest-neighbor clustering. Some points belonging to the same trunk are inevitably grouped into separate clusters due to occlusions and error in the reconstruction. To cope with this, TreeTool groups same-trunk clusters together. We add an extra clustering step using density based clustering applications with noise (DBSCAN)~\cite{schubert2017dbscan} which addresses the case where points from different trunks are grouped together. This is especially prevalent for resprouting trees like redwoods, which commonly grow with conjoined trunk bases.

The last stage involves modeling the segmented and filtered trees to estimate diameter and location. Tree clusters are vertically cropped such that the remaining clusters represent the trunks between 1.0 and 1.6 m above the modeled ground surface. DBH is estimated by taking the maximum diameter reported between cylinder and ellipse fitting methods. Random sample consensus (RANSAC) fits a cylinder to the cropped trunk cluster. An additional ellipse model is generated using least-squares on a 2D projection of the points.

\subsubsection{Convex hull modeling approach}
The use of RANSAC for modeling trees as cylinders to estimate DBH is common in the literature~\cite{pierzchala2018mapping, freissmuth2024online}. An advantage of this approach is that DBH can be extrapolated from partially represented tree trunks, a common occurrence since optimal scene coverage is often not possible in complex forest terrain. A downside of this method is that for well-represented trunks, a cylinder model is prone to underfitting the true trunk diameter as seen in Figure~\ref{fig:convex_hulls}. This is even more prevalent for species with deeply furrowed, irregular bark texture. Another limitation of this method is the inability to model irregular, bowed trunk shapes. 

We propose a modeling approach which considers tree clouds as a stacks of convex-hull slices as seen in Figure~\ref{fig:treetool}. The trunk is vertically partitioned into 20 cm thick slices. Each slice is extracted, and oriented to be colinear with the z-axis. A 2D xy projection of the points is used to fit a convex-hull around the surface of the cloud; this emulates manual DBH measurement via girth tape. To deal with noisy points, we introduce a layer of DBSCAN that removes low-density regions. DBH is estimated by considering the slice at 1.3 m above the ground. We take the maximum value across LS, RANSAC, and convex-hull methods as the final diameter to account for partial trunk cases. DBSCAN parameters control the maximum distance $\epsilon$ of points to be considered in a neighborhood, and the minimum point count (minPts) within that region to be considered a dense region. We found an $\epsilon$ range between 1 and 3 cm to have good outlier rejection on 2D trunk clouds. The minPts parameter is dependent on the 2D surface density $\rho_c [\frac{\text{points}}{\text{m}^2}]$ of the trunk cloud; we observed successful filtering in the range of 5-40 points.

\subsection{Study area and data collection}
To validate the precision and accuracy of the proposed NeRF-derived convex-hull DBH method, we conducted an experiment in the Forest Ecology Research Plot (FERP)~\cite{gilbert2024three}, a globally recognized ForestGEO~\cite{davies2021forestgeo} site in the Santa Cruz mountains along the central coast of California, USA. This plot spans 16 ha with over 51,000 recorded stem locations and DBH measurements. The forest census is repeated on a 5-year cycle.

The FERP is partitioned into 400 20$\times$20 meter subplots denoted by E\textbf{x}\_N\textbf{y} where \textbf{x} and \textbf{y} are the distance in meters from the SW corner of the FERP (37.012416, -122.074833) to the SW corner of the subplot. This study considered sections of forest in subplots E340\_N360 and E340\_N380. The data collection effort was accomplished over two visits and spans two datasets. LiDAR and IMU data were recorded at 10 Hz and 500 Hz respectively. NeRF training data was collected using an iPhone 14 camera (1920$\times$1440) and NeRFCapture~\cite{nerfcapture}. Training via nerfstudio~\cite{nerfstudio} lasted 300K iterations and took 15 minutes for both datasets. As a reference technique, DBH was taken manually via girth tape by a trained research assistant.

\begin{figure*}[!ht]
    \centering
    \includegraphics[width=0.9\textwidth]{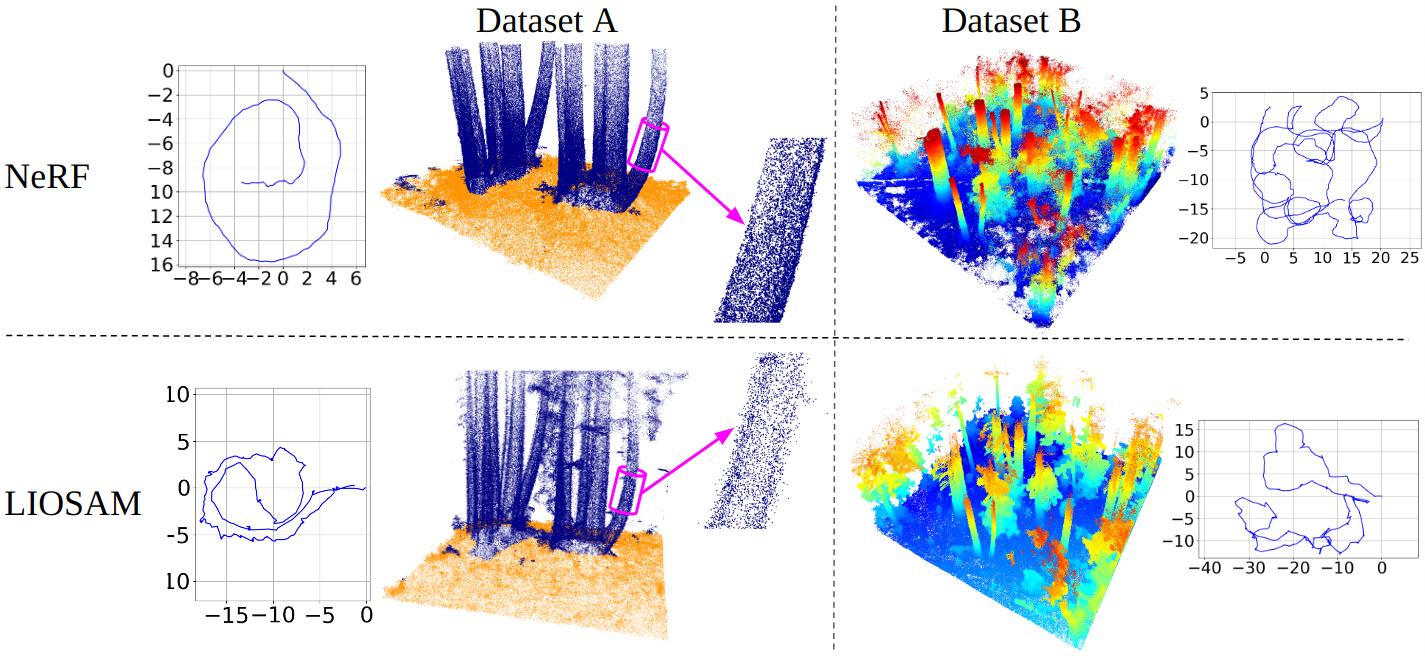}
    \caption{Forest reconstructions produced by SLAM (bottom row) and NeRF (top row) methods of both datasets. Adjacent plots are data collection trajectories for each reconstruction. The figure also compares a zoomed-in section of a tree trunk. The NeRF reconstruction is 4$\times$ denser and is of higher surface quality.}
    \label{fig:reconstructions}
\end{figure*}

\begin{table*}[!ht]
        \begin{center}
        \caption{Study area parameters, quantity of recorded data, and a comparison of reconstruction density and fieldwork duration between NeRF and SLAM approaches. Duration includes field-validation of recorded data, not post processing.}
        \renewcommand{\arraystretch}{1.2}
            \begin{tabular}{ l|c c c c|c c c|c c} 
                & & & & & \multicolumn{3}{c|}{Fieldwork duration} & \multicolumn{2}{c}{Point count}\\
                 Dataset &  \textbf{Area (m$^2$)} & \textbf{Tree count} & \textbf{LiDAR frames} & \textbf{Images} & \textbf{NeRF} & \textbf{SLAM} & \textbf{Manual} & \textbf{NeRF} & \textbf{SLAM} \\ \hline
                A & 140 & 11 & 2,172  & 166 & 5 min & 30 min & 45 min & 2.69M & 704K \\
                B & 400 & 9 & 7,498 & 847 & 8 min & 40 min & 52 min & 26.87M & 7.83M
                \label{tab:visual_results}
            \end{tabular}
        \end{center}
\end{table*}

\renewcommand{\arraystretch}{1.2} 
\begin{table*}[!ht]
\centering
\caption{DBH estimation accuracy between our proposed convex-hull approach and RANSAC for NeRF and SLAM reconstructions.}
\begin{tabular}{l|cc|cc|cc|cc}
\hline
& \multicolumn{4}{c|}{Dataset A} & \multicolumn{4}{c}{Dataset B}  \\
& \multicolumn{2}{c}{\textbf{NeRF}} & \multicolumn{2}{c|}{\textbf{SLAM}} & \multicolumn{2}{c}{\textbf{NeRF}} & \multicolumn{2}{c}{\textbf{SLAM}} \\ \hline
\textbf{Metric} & Convex-hull & RANSAC & Convex-hull & RANSAC & Convex-hull & RANSAC & Convex-hull & RANSAC \\ \hline
Bias & -0.28 cm & -4.35 cm & -1.35 cm & -6.89 cm & -0.86 cm & -4.59 cm & 1.56 cm & -3.69 cm \\
RMSE & 1.26 cm & 4.96 cm & 2.32 cm & 7.12 cm & 2.09 cm & 5.28 cm & 1.93 cm & 4.53 cm \\
Std & 1.29 cm & 2.49 cm & 1.97 cm & 1.86 cm & 2.02 cm & 2.77 cm & 0.93 cm & 2.88 cm 
\label{tab:results}
\end{tabular}
\end{table*}

\subsubsection{Dataset A}
In the first dataset, the robot was teleoperated around a cluster of 11 coast redwood trees (\textit{Sequoia sempervirens}) to generate a SLAM reconstruction. The robot was also navigated through an opening between the trees to recover additional occluded trunk and ground geometry within the cluster. NeRF training data was taken by traversing the tree cluster in a similar fashion. The trees had 360$^\circ$ coverage in the training data since nearby terrain afforded easier maneuvering. 

\subsubsection{Dataset B}
The second dataset is larger by area, spanning the entire E340\_N360 subplot which included more challenging terrain and foliage occlusions. This area consisted of 6 coast redwood and 3 Douglas-fir (\textit{Pseudotsuga menziesii}) trees. The robot's obstacle-avoidance mode enabled maneuvering in complex terrain, but at a significantly reduced pace (Table~\ref{tab:visual_results}). Stems less than 8 cm in diameter where not considered in this study, as robust DBH estimation was unstable in this size range.

\section{Results}
\label{sec:results}
We assess the accuracy of DBH estimation for each dataset independently and combined, using bias, which gives an idea of over/under estimation-trends), root mean squared error (RMSE), and standard deviation as these are commonly referenced metrics in this domain. These are defined as:
\begin{gather}
    \text{Bias}=\frac{1}{n}\sum_{i=1}^n(y_i-y_{ri})\\
    \text{RMSE}=\sqrt{\frac{\sum(y_i-y_{ri})^2}{n}}
\end{gather}
where $y_i$ and $y_{ri}$ are the estimated and reference diameters across $n$ estimations. Our method consistently outperformed RANSAC/LS, which was prone to under-fitting (Table~\ref{tab:results}; Figure~\ref{fig:convex_hulls}). NeRF reconstruction with convex-hull trunk modeling yielded a low of 2.09 cm RMSE and a best case 1.26~cm RMSE with an average of 1.68 cm across both datasets (20 trees). In general, our approach outperformed RANSAC cylinder modeling by 3-4$\times$. In collecting data in the forest, NeRF outperformed LIOSAM (2.15 cm RMSE), with a 5$\times$ faster time compared to LIOSAM and 10$\times$ compared to manual measurement. 

The NeRF reconstructions were consistently 3-4$\times$ more dense than the SLAM reconstructions (Table~\ref{tab:visual_results}). The sparsity associated with LIOSAM is due to the nature of points being registered by laser pulse returns that have a resolution of 262k points (128$\times$2048) per LiDAR frame, a physical limitation of the hardware. NeRFs are capable of higher density reconstruction since the geometry is rendered by sampling the learned color-ray space and filtering out low-density sections to only represent surfaces. This increased point density is a major advantage of NeRFs.

\section{Discussion}
\label{sec:discussion}
The forest environment consists of harsh lighting conditions that add challenges to the use of photometric methods such as NeRFs. The dark understory created by dense forest canopy requires appropriate exposure control; long exposure times can lead to blurry images that are unusable for reconstruction purposes. A potential solution is offered by RawNeRF~\cite{mildenhall2022nerf} which enables reconstruction in near-darkness environments. 

The NeRF method's impressive speed-up of fieldwork time comes with the challenge  of reconstructing and modeling smaller stems and complex foliage. Without optimal camera coverage, this geometry is poorly represented by the NeRF. Additionally, the filtering methods used by TreeTool need to be developed to support smaller stems and foliage. One potential avenue for improved clustering and filtering performance is to use the color of points provided by NeRF representation to aid in complex branch and foliage segmentation. 

We show that convex-hull modeling is an improvement over cylinder approaches for measuring tree diameters when well-represented tree clouds are available. A potential solution for partial clouds could be to interpolate the cross-section from the set of stacked convex-hulls at known spacing along its height. This same slice-based modeling could be used to extract additional science measurements beyond DBH.

Parameter selection is still a semi-manual process for TreeTool. Not all parameters can be derived from density (e.g., terrain morphology dictates ground and trunk segmentation parameters). Robust, automatic-parameterization can enable real-time DBH estimation in various rugged settings; further work is needed to understand the relationship between the parameter selection process and the tree-species composition of the environment. 

\section{Conclusion}
\label{sec:conclusion}
In summary, we present a field study exploring the benefits of NeRFs for DBH estimation in mixed-evergreen redwood forest. We consider an MLS comparison using LiDAR-inertial SLAM hosted on a quadruped robot. In addition, we propose a convex-hull DBH modeling technique which considerably outperformed common cylinder-fitting approaches by 3-4$\times$. In a small-scale experiment, NeRF reconstructions made using mobile phone data outperformed SLAM in terms of DBH estimation accuracy (1.68 cm RMSE), at a 20$\times$ cost reduction and 5$\times$ less time. Using NeRF and an improved convex-hull tree modeling approach outperforms several published forest mapping papers~\cite{pierzchala2018mapping,freissmuth2024online,liang2013, huang2024}. 

While additional development is needed for autonomous ecological assessment in wilderness settings, this paper motivates the ability for rapid forest data collection using commodity mobile phone hardware. This drastic increase in accessibility has the potential of furthering community engagement, and increasing the volume of globally mapped forest terrain. 

\section*{Acknowledgments}
We would like to thank Morgan Masters for assisting in field data collection and providing feedback during the writing of this paper. This work was funded by the Office of Naval Research (grant number N00014-22-1-2290) and the UCSC Center for Information Technology Research in the Interest of Society (CITRIS).

\ifCLASSOPTIONcaptionsoff
  \newpage
\fi



%
\bibliographystyle{IEEEbib}
\bibliography{strings,refs}

%

\begin{IEEEbiographynophoto}{Adam Korycki}
is a researcher of Applied Robotics at UC Santa Cruz. He received a B.S. in Computer Engineering and a M.S. in Electrical and Computer Engineering at UC Santa Cruz. His research invovled applications of robotics and Neural Radiance Fields (NeRFs) for forest forest mapping, with an emphasis on environmental data collection and analysis. His research interests are in field robotics, computer vision, and 3D reconstruction with a focus on environmental and sustainability applications.
\end{IEEEbiographynophoto}

\begin{IEEEbiographynophoto}{Cory Yeaton}
Cory Yeaton graduated from UC Santa Cruz with a B.S. in Ecology and Evolutionary Biology in 2024. His studies primarily focused on botany, plant ecology and ecological restoration techniques. During his final year of undergraduate work, he completed a thesis evaluating the efficiency of various forest mapping procedures in the Santa Cruz Mountains. Cory currently works as a seasonal ecology research technician and a habitat restoration assistant at Musick Creek in Shaver Lake, California. He is passionate about ethical stewardship of wild places and expanding his botanical knowledge with the goal of protecting and restoring the natural world.

\end{IEEEbiographynophoto}


\begin{IEEEbiography}[{\includegraphics[width=1in,height=1.25in,clip,keepaspectratio]{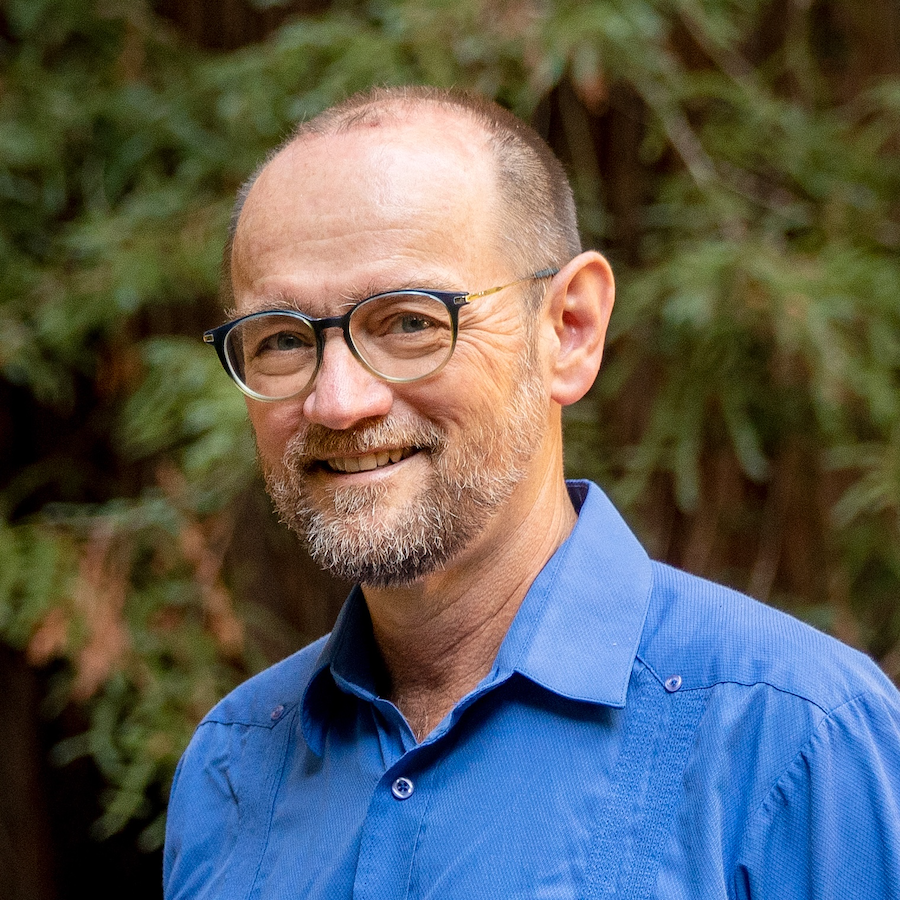}}]{Gregory S. Gilbert}
 received a B.Sc. degree in Environmental and Forest Biology from the SUNY College of Environmental Science and Forestry, and M.Sc. and Ph.D. degrees in Plant Pathology from the University of Wisconsin - Madison.  He is Professor of Environmental Studies at the University of California Santa Cruz, Director of the UCSC Forest Ecology Research Plot, and Research Associate at the Smithsonian Tropical Research Institute in Panama. His research interests are in forest ecology, the evolutionary ecology of plant diseases, and applied phylogenetic ecology.
\end{IEEEbiography}

\begin{IEEEbiography}
[{\includegraphics[width=1in,height=1.25in,clip,keepaspectratio]{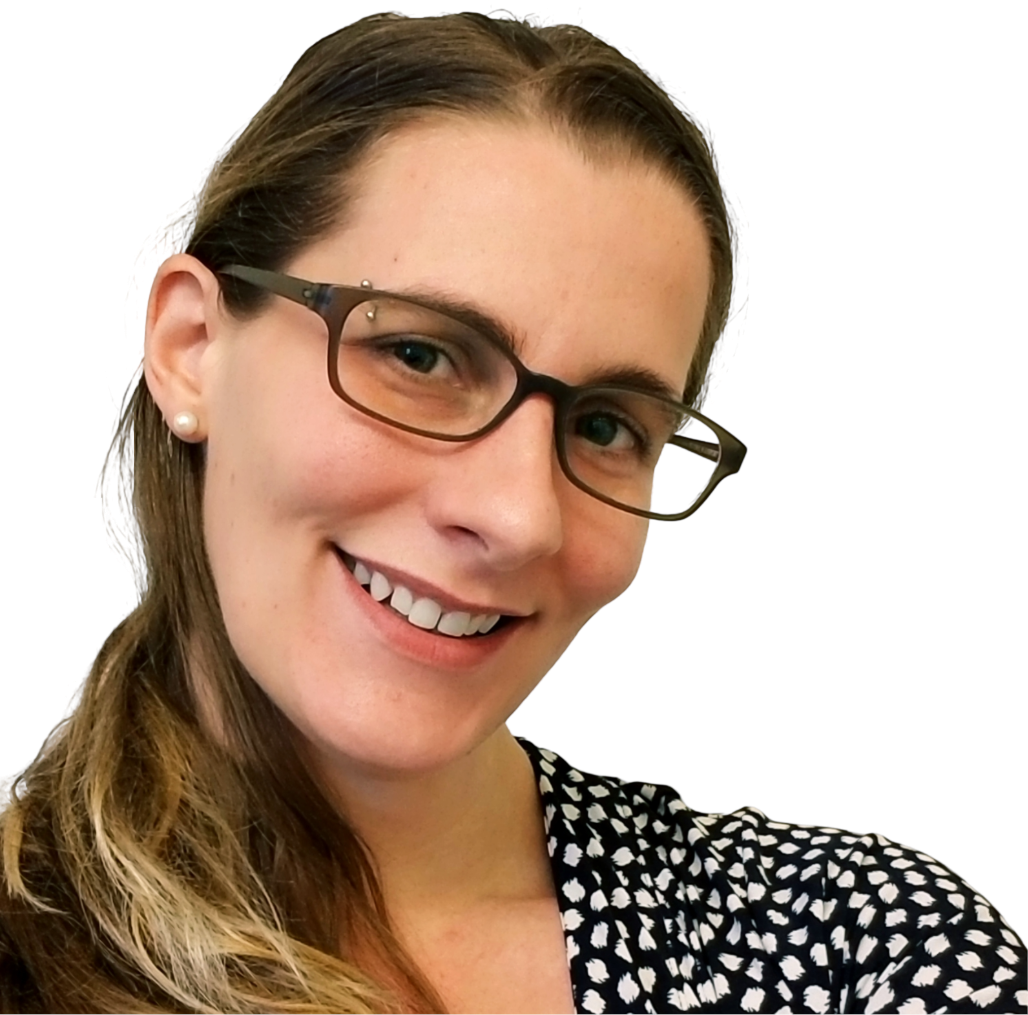}}]{Colleen Josephson}
 is an Assistant Professor of Electrical and Computer Engineering and AES (Agricultural Experiment Station) Agronomist at UC Santa Cruz. Her research focuses on wireless sensing systems to enable and improve sustainable practices. She is an associate editor for IEEE Transactions on AgriFood Electronics. Colleen completed her PhD in Electrical Engineering at Stanford University, and received S.B and MEng degrees from the Massachusetts Institute of Technology. 

\end{IEEEbiography}

\begin{IEEEbiography}
[{\includegraphics[width=1in,height=1.25in,clip,keepaspectratio]{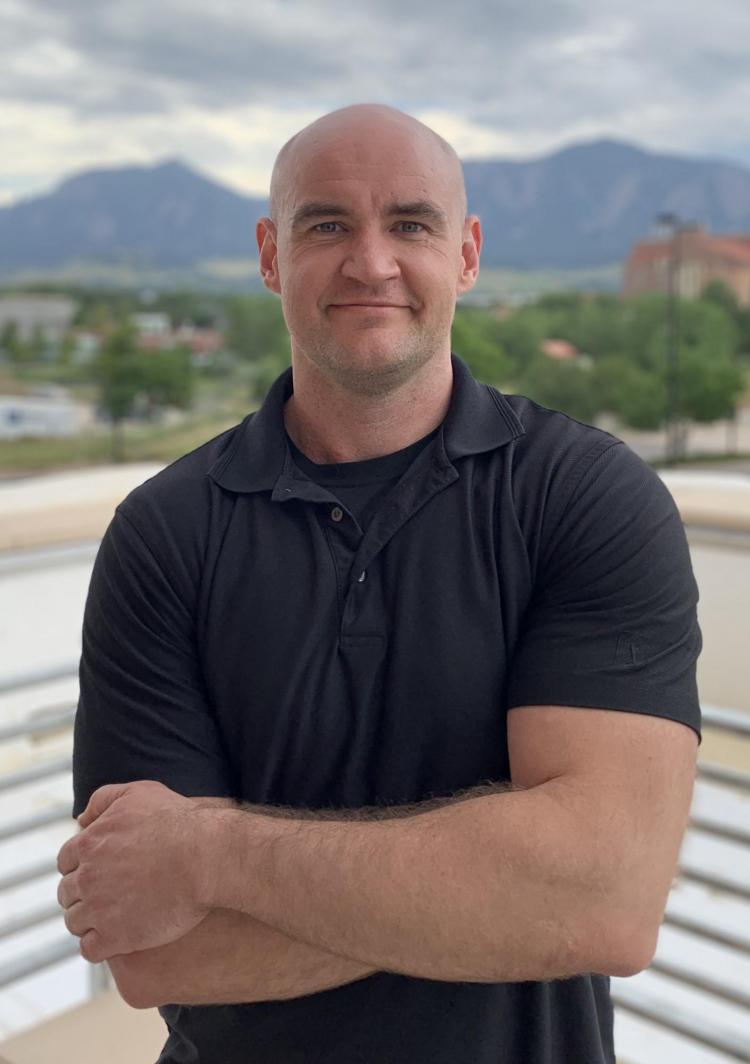}}]{Steve McGuire} is an Assistant Professor of Electrical and Computer Engieneering at UC Santa Cruz. His research develops techniques in field robotics, including perception, navigation, and planning, to better explore real-world challenging environments. His current projects explore how to leverage advanced robotics in ecology and agriculture to expand scientific inquiry. Steve completed his PhD in Aerospace Engineering Sciences from the University of Colorado Boulder and has been supported by DARPA, ONR, NASA, USDA-NIFA, as well as industry. 

\end{IEEEbiography}




\end{document}